\documentclass{article}

\usepackage{PRIMEarxiv}

\usepackage[utf8]{inputenc} 
\usepackage[T1]{fontenc}    
\usepackage{hyperref}       
\usepackage{url}            
\usepackage{booktabs}       
\usepackage{amsfonts}       
\usepackage{nicefrac}       
\usepackage{microtype}      
\usepackage{lipsum}
\usepackage{graphicx}
\graphicspath{{media/}}     

\usepackage{indentfirst}
\usepackage{amsmath}
\usepackage{textcomp}
\usepackage{xcolor}
\usepackage{overpic} 
\usepackage{amsmath,bm}
\usepackage{multirow}
\usepackage{array}
\usepackage{tabularx}
\usepackage{dcolumn}
\usepackage{color}

\newtheorem{theorem}{Theorem}
\newtheorem{lemma}{Lemma}

\usepackage{algorithm}
\usepackage{algorithmic}
  
\title{Automated Proof of Polynomial Inequalities via Reinforcement Learning
}
\author{Banglong\;Liu$^{1}$,Niuniu\;Qi$^{1}$,Xia\;Zeng$^{2}$,Lydia\;Dehbi$^{1}$,Zhengfeng\;Yang$^{1}$\thanks{Corresponding author.} \\
$^{1}$East China Normal University,$^{2}$Southwest University\\
\texttt{ \{blliu,jnaite\}@stu.ecnu.edu.cn},\texttt{ xzeng0712@swu.edu.cn,} \\
\texttt{ \{dehbilydia,zfyang\}@sei.ecnu.edu.cn} 
}

\begin{document}
\maketitle

\begin{abstract}
Polynomial inequality proving is fundamental to many mathematical disciplines and finds wide applications in diverse fields. 
Current traditional algebraic methods are based on searching for a polynomial positive definite representation over a set of basis. However, these methods are limited by truncation degree.
To address this issue, this paper proposes an approach based on reinforcement learning to find a {Krivine-basis} representation for proving polynomial inequalities.
Specifically, we formulate the inequality proving problem as a linear programming (LP) problem and encode it as a basis selection problem using reinforcement learning (RL), achieving a non-negative {Krivine basis}. 
Moreover, a fast multivariate polynomial multiplication method based on Fast Fourier Transform (FFT) is employed to enhance the efficiency of action space search. 
Furthermore, we have implemented a tool called {APPIRL} (Automated Proof of Polynomial Inequalities via Reinforcement Learning).
Experimental evaluation on benchmark problems demonstrates the feasibility and effectiveness of our approach. In addition, {APPIRL} has been successfully applied to solve the maximum stable set problem.
\end{abstract}
\section{Introduction}
\label{sec:intro}

Polynomial inequalities are fundamental in mathematics and arise in various fields such as optimization \cite{BV2014}, control theory \cite{parrilo2000structured}, algebraic geometry \cite{parrilo2004inequality} and combinatorics \cite{marechal2015polyhedral}.
As problems in these fields increase in complexity, the demand for efficient and scalable methods to prove polynomial inequalities has grown. 
This has led to the development of various computational approaches, each offering different trade-offs in terms of completeness, efficiency, and applicability, particularly for high-degree polynomials or systems with many variables.

Proving polynomial inequalities involves demonstrating that a given polynomial is non-negative over a specified domain. 
Traditional algebraic methods are grounded in solving for a positive definite representation of the target polynomial, and typically rely on static proof systems that define inference rules enabling the derivation of new polynomial inequalities from existing ones. 
Specifically, Kaltofen et al. \cite{KALTOFEN20121} employed sums-of-squares (SOS) and Gauss-Newton iteration methods to derive exact lower bounds for polynomial systems. 
Yang et al. \cite{Yang2023} applied Bertini and MMCRSolver to approximate real solutions and used AINLSS to precisely verify the existence of solutions for polynomial systems.
The Sherali-Adams framework \cite{SheraliA90} and the Lovász-Schrijver method \cite{lovasz1991cones} proposed sequences of linear and semidefinite relaxations that tightened the approximation of the original problem by incorporating additional constraints. 
Lasserre’s hierarchy \cite{lasserre2002} further refined this approach by systematically constructing moment and SOS relaxations through SDP, providing a systematic sequence of approximations that converge to the exact solution or offer highly accurate bounds. 
However, the aforementioned traditional methods are constrained by the choice of truncation degree $D$. Setting $D$ too high can result in prohibitively large time complexity, rendering the problem intractable. Conversely, if $D$ is set too low, the expressiveness of the solution may be restricted, or no solution may exist, thus making verification infeasible. 
As a result, achieving significant breakthroughs with such methods remains challenging.

One popular approach is learning-based dynamic proofs. 
This method leverages artificial intelligence (AI) to dynamically explore proofs for polynomial positive definite expressions, circumventing the limitations imposed by the choice of truncation order in solving the problem.
Grigoriev et al. \cite{grigoriev2002} demonstrated that certain problems requiring large static proofs could often be resolved more efficiently using compact dynamic proofs. 
More recently, machine learning techniques have been increasingly integrated into theorem proving to enhance the efficiency and accuracy of proof generation \cite{huang2018,kaliszyk2018,bansal2019}. 
These techniques have focused on optimizing the selection of inference rules or tactics during the proof process. 
Fawzi et al. \cite{FawziMFF19} proposed a dynamic proof system, guided by a neural network, for addressing the maximum stable set problem. This system is capable of learning to select appropriate inference rules during the proof process.

This paper proposes a novel approach to proving polynomial inequalities based on reinforcement learning (RL). Building on the Positivstellensatz theorem \cite{krivine1964}, we formulate the inequality proof problem as finding a non-negative representation in the {\em Krivine-basis} form. 
The optimal base selection strategy is then trained using reinforcement learning.
During the training process, it incrementally and effectively adjusts the optimal basis to precisely update the basis library for the polynomial positive definite representation. 
This results in a lightweight polynomial expression that fully captures the sparsity of the polynomial representation and avoids the coefficient inflation problem associated with conventional polynomial parameter expressions. 
Additionally, for the polynomial multiplication operations involved in the intermediate solution process, we propose a fast polynomial multiplication expansion method based on Fast Fourier Transform (FFT), significantly improving computational efficiency.
The main contributions of this paper are summarized as follows:
\begin{itemize}
    \item We formulate the polynomial inequality proving problem as a construction problem of {\em Krivine-basis} representation, and develop a reinforcement learning-based framework that incrementally selects effective bases to obtain a non-negative representation.

   \item  We propose a fast multivariate polynomial multiplication method based on multivariate affine transformations and Fast Fourier Transform (FFT), accelerating the computation of basis.

    \item We conduct a comprehensive experimental evaluation on a set of benchmarks, demonstrating the efficiency and effectiveness of the proposed approach. Our method is successfully applied to the maximum stable set problem, showing its practical applicability.
\end{itemize}

\section{Preliminaries}
\label{sec:Preliminaries}

{\bf [Notations.]} 
Let $\mathbb{R}[x] = \mathbb{R}[x_1, \ldots, x_n]$ denote the multivariate polynomials ring in $n$ variables with real coefficients. ${\bf x}^{\alpha}=x_1^{\alpha_1}x_2^{\alpha_2}\cdots x_n^{\alpha_n}$ represents a monomial, where $\alpha= (\alpha_1,\ldots,\alpha_n) \in \mathbb{N}^n$. 
The degree of ${\bf x}^{\alpha}$ is $|\alpha|=\sum_{i=1}^n\alpha_i$. 

In this paper, we address the problem of proving polynomial inequalities on a specific type of semi-algebraic set, namely, the hyperrectangle:
\begin{equation}\label{problemNON}
f({\bf x})\geq 0, \quad \forall {\bf x}\in\mathcal {S},
\end{equation}
where $f({\bf x})$ is a polynomial in the polynomial ring $\mathbb{R}[{\bf x}]$, and 
$$\mathcal {S}=\{{\bf x}\in \mathbb{R}^n|x_i\in [a_i,b_i],i=1,\ldots,n\}.$$ 

Without loss of generality, we can consider continuous functions $f({\bf k})$ in the unit hypercube $[0, 1]^n$ through a change of variables:
$k_i = (x_i - a_i)/(b_i - a_i)$.
This substitution maps $x_i\in [a_i, b_i]$ to $k_i\in [0, 1]$ preserving the maximum norm of any function. 
Consequently, unless explicitly stated otherwise, we focus solely on the non-negativity of $f({\bf x})\geq 0$ on the unit hypercube $\mathcal{S} = \{{\bf x}\in [0, 1]^n\}$.

The non-negativity of the polynomial $f({\bf x})$ on the unit hypercube ${\bf x}\in [0, 1]^n$ can be proven using the inference rules of the proof system \cite{LovaszS91} and the Sherali-Adams framework \cite{SheraliA90}. These rules include:

\begin{equation}\label{proveruel}
\begin{array}{ll}
{g\geq 0}\rightarrow{x_ig\geq 0},\; {(1-x_i)g\geq 0 }\\
{h_i\geq 0}\rightarrow{\sum_i\lambda_ih_i\geq 0}, \forall \lambda_i\geq 0,
\end{array}
\end{equation}
where $g \geq 0$ and $h_i \geq 0$ represent the given inequalities, which can be interpreted as some non-negativity conditions.
The basic inference rules can be combined to obtain a function in the form
${\bf x}^{\alpha}(1-{\bf x})^{\beta}$, where $\alpha,\beta\in \mathbb{N}^n$ and $(1-{\bf x})^\beta=(1-x_1)^{\beta_1}\cdots(1-x_n)^{\beta_n}$. 
This form guarantees the non-negativity of the function for all ${\bf x}\in [0,1]^n$. Therefore, expressing a polynomial in this form serves as a proof of its non-negativity on the unit hypercube. We now state the Positivstellensatz theorem.

\begin{theorem}\label{TheoremPosi}
(Positivstellensatz \cite{krivine1964}).
Suppose $f$ is a polynomial that satisfies $f({\bf x}) > 0$ for all ${\bf x}\in [0,1]^n$. Then there exists an integer $l$ and non-negative scalars $\lambda_{\alpha,\beta}\geq 0$ such that:
\begin{equation}\label{theorempositivst}
f({\bf x})=\sum_{|\alpha|+|\beta|\leq l}\lambda_{\alpha,\beta}{\bf x}^{\alpha}(1-{\bf x})^{\beta}.
\end{equation}
\end{theorem}

According to Theorem \ref{TheoremPosi}, the existence of the expression (\ref{theorempositivst}) provides a sufficient condition for the non-negativity of the polynomial $f({\bf x})$ on $\mathcal {S}$. 
However,  the exact determination of $l$ poses a challenge in constructing the representation (\ref{theorempositivst}). 
To address this, we introduce a degree bound $D$ to relax Eq.(\ref{theorempositivst}) and correspondingly select the basis ${\bf x}^{\alpha}(1-{\bf x})^{\beta}$. 
This approach allows us to construct a sufficient condition for $f({\bf x})\geq 0$ on $\mathcal {S}$.

\begin{lemma}\label{lemma}
Let $\mathcal{S}$ be a unit hypercube. 
If there exists a positive integer $D$ such that $f({\bf x})$ can be represented as 
\begin{equation}\label{theorem:fun}
f({\bf x})=\sum_{|\alpha|+|\beta|\leq D}\lambda_{\alpha,\beta}{\bf x}^{\alpha}(1-{\bf x})^{\beta},
\end{equation}
then $f({\bf x}) \geq 0$ holds on $\mathcal{S}$.
\end{lemma}

We refer to the expression in Eq.(\ref{theorem:fun}) as the {\em Krivine-basis} representation.
If such a $D$ can be determined, then the inequality $f({\bf x})\geq 0$ can be proved on the unit hypercube. Therefore, in the following sections, we will focus on the construction of the {\em Krivine-basis} representation.
\section{LP Problem Transformation}
\label{sec:trans}

To better illustrate the connection between the expression (\ref{theorem:fun}) and the proof objective, namely, the non-negative representation in the {\em Krivine-basis} form, we can reformulate the problem as an optimization problem.
Specifically, the problem (\ref{problemNON}) can be transformed into the following problem:
\begin{equation}\label{opt:orig}
\left.\begin{array}{l}
{\rm max}\quad \gamma \\
s.t. \quad f({\bf x})-\gamma =\sum\limits_{|\alpha|+|\beta|\leq D}\lambda_{\alpha,\beta}{\bf x}^{\alpha}(1-{\bf x})^{\beta}, \\
\quad\quad\;\lambda_{\alpha, \beta}\geq 0.
\end{array}\right\}
\end{equation}

Notably, the equality constraint in (\ref{opt:orig}) represents a functional equality between two polynomials, which is equivalent to the equality of their corresponding polynomial coefficients. Thus, we can rewrite the equality constraint in problem (\ref{opt:orig}) as a linear problem with respect to the variables $\gamma$ and $\lambda$.

Let $[{\bf x}]_d$ denote the vector of monomials with degrees at most $d$, ordered in graded lexicographic order, i.e.,
$${\bf v}=[1,x_1,x_2,\ldots,x_n,x_1^2,x_1x_2,\ldots,x_{n-1}x_n^{d-1}, x_n^d]^{T},$$
where the dimension of $[{\bf x}]_d$ is $t=\dim([{\bf x}]_d)=\binom{n+d}{n}$.
Thus, a polynomial $f({\bf x})$ can be expressed as 
$$f({\bf x})=\sum_{|\alpha|\leq d}c_{\alpha}{\bf x}^{\alpha}={\bf c}^{T}{\bf v},$$ 
where ${\bf c} = [c_0, c_1, \ldots, c_t]^{T}$ is the coefficient vector.

Similarly, the right-hand side of the equality in problem (\ref{opt:orig}) can be expressed linearly. 
By extracting the corresponding coefficient equalities, we obtain the following linear programming problem:
\begin{equation}\label{opt:matrix}
\left.\begin{array}{ll}
{\rm max}\quad \; \gamma \\
s.t. \quad\;\; A\cdot [{\bf {\lambda}}, \gamma]^T = {\bf c},\\
\qquad \quad {\bf {\lambda}} \succeq 0,
\end{array}\right\}
\end{equation}
where ${\bf {\lambda}}$ is a vector corresponding to $\lambda_{\alpha,\beta}$, and $A$ is a coefficient matrix of $\gamma$ and $\lambda$, obtained by lexicographically sorting $[{\bf x}]_d$ from the equation (\ref{opt:orig}).
The linear programming problem (\ref{opt:matrix}) can be solved using algorithms such as the interior-point method \cite{BV2014} or the simplex method \cite{Dantzig1951}.

According to Theorem \ref{TheoremPosi}, if $f({\bf x})$ is non-negative on the unit hypercube $[0,1]^n$, then there exists a {\em Krivine-basis} representation such that the optimal solution $\gamma^{*}$ of the problem (\ref{opt:matrix}) satisfies $ \gamma^{*} \geq 0$. Leveraging this property, we present the following theorem.

\begin{theorem}\label{Theorem3}
Given a polynomial $f({\bf x}) \in \mathbb{R}[{\bf x}]$, if there exists a positive integer $D$ such that the optimal solution $\gamma^{*}\geq 0$ of problem (\ref{opt:orig}), then $f({\bf x})$ is non-negative on the unit hypercube.
\end{theorem}

\noindent
{\em Proof.}  
If there exists a positive integer $D$ such that the solution $\gamma^{*}$ of problem (\ref{opt:orig}) is non-negative, then we can express $f({\bf x})$ as $\gamma^{*} + \sum_{|\alpha|+|\beta|\leq D}\lambda_{\alpha,\beta}{\bf x}^{\alpha}(1-{\bf x})^{\beta} \geq 0$. This implies that $f({\bf x})$ is non-negative on $[0,1]^{n}$. Thus, the theorem \ref{Theorem3} is proved.  $\hfill\square$ 

If the optimal solution $\gamma^{*}$ obtained by solving the LP problem (\ref{opt:matrix}) satisfies $\gamma^{*}\geq 0$, then we can conclude that the polynomial $f({\bf x})$ is non-negative on the unit hypercube ${\bf x}\in [0,1]^n$. Otherwise, the inequality cannot be proven.
A key property is that as the degree bound $D$ increases, the expressive power of the model increases, resulting in an increase in the value of $\gamma^{*}$.
Thus, as $D$ approaches infinity, there exists $\gamma^{*} \geq 0$ for problem (\ref{opt:matrix}) such that $f({\bf x}) \geq 0$ holds. However, this leads to an explosion in dimension, which makes the problem computationally intractable.
When $D$ is small, the number of rows in the matrix $A$ may exceed the dimension of the variable ${\bf y}$, rendering problem (\ref{opt:matrix}) infeasible.
Therefore, traditional methods relying on a fixed degree bound $D$ are often unable to overcome these challenges.

To address this issue, we propose employing a reinforcement learning (RL) approach to progressively select the bases ${\bf x}^{\alpha}(1-{\bf x})^{\beta}$, thereby constructing a {\em Krivine-basis} representation to prove that the polynomial inequality holds on $\mathcal{S}$.
This approach enables the degree to increase gradually, thus avoiding the need to exhaustively explore the degree bound $D$ and reducing unnecessary computations.
\section{Dynamic Proof via Reinforcement Learning}
\label{prove}

\begin{figure*}[ht]
 \centering 
 \includegraphics[width=1\textwidth]{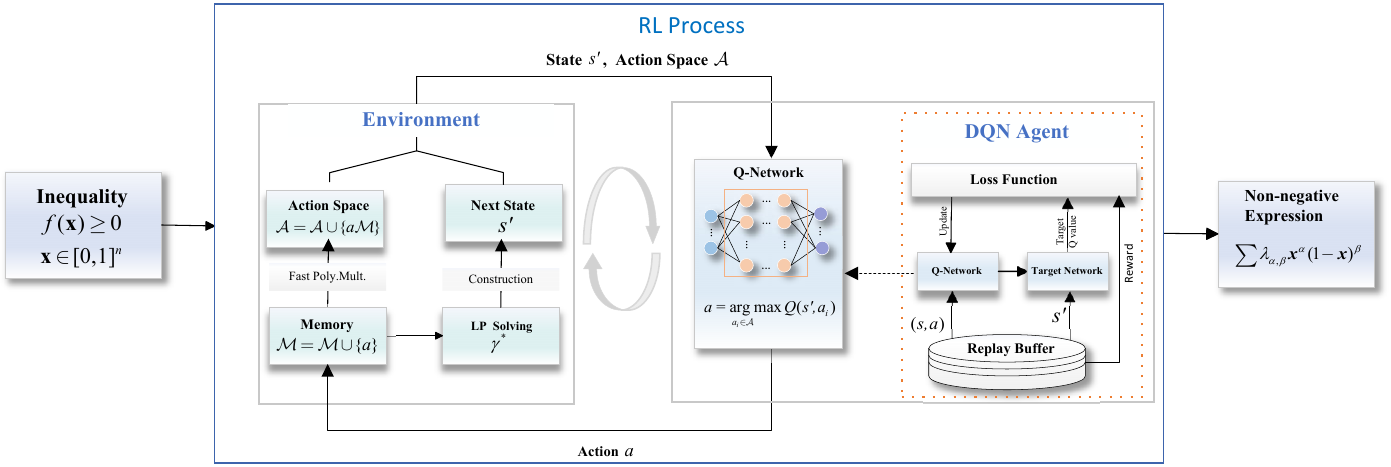}
 \caption{The framework of automated proof of polynomial inequalities based on RL} 
 \label{figure:frame}
 \vspace{-0.4cm}
\end{figure*}

In this section, we explore the application of reinforcement learning (RL) to address the problem outlined in (\ref{opt:orig}). 
The objective is to train an agent to progressively select the basis ${\bf x}^{\alpha}(1-{\bf x})^{\beta}$ throughout the proof process, ultimately obtaining a non-negative representation.
To achieve this, we propose an automated framework for proving polynomial inequalities based on the Deep Q-Network (DQN) \cite{2013playing}, as shown in Fig. \ref{figure:frame}.

As illustrated in Fig.\ref{figure:frame}, our method consists of two main components: {\em Environment Construction} and {\em DQN Training}. 
In the {\em Environment Construction} component, the memory $\mathcal{M}$ is updated with the selected action $a$.
Since the objective is to find a non-negative representation in the {\em Krivine-basis} form, both the action $a$ and memory $\mathcal{M}$ are in the form of bases ${\bf x}^{\alpha}(1-{\bf x})^{\beta}$. 
We then solve the LP problem (\ref{opt:matrix}) to obtain the optimal solution $\gamma^{*}$, which is used to update the next state $s'$ and generate an immediate reward $r$.
Furthermore, we implement a fast multivariate polynomial multiplication to update the action space $\mathcal{A}$, thereby enhancing computational efficiency.
In the {\em DQN Training} component, data is sampled from the replay buffer $R$ to train the Q-network. This network guides the selection of a basis $a$ from the basis space $\mathcal{A}$, based on the immediate reward $r$, and feeds it back to the environment. 
If a non-negative representation $\sum\lambda_{\alpha,\beta}{\bf x}^{\alpha}(1-{\bf x})^{\beta}$ is found, then the inequality proof is successfully completed; otherwise, the proof cannot be established.
In the subsequent sections, we will elaborate on the construction process of the non-negative representation.

\subsection{Environment Construction}
We model the process of finding a non-negative representation of a polynomial inequality as an interaction between the agent and the environment, formalized as a Markov Decision Process (MDP) \cite{bellman1966dynamic}. This interaction generates a sequence of states, basis functions, and observed rewards. The decision-making process of the agent at time step $t$ is defined by a triplet $(s_t,{\mathcal {A}}_t, r_t)$, consisting of the following key elements:
 \begin{itemize}
\item $s_t$ represents the current state, composed of the proximity value $\gamma$ of the proof objective and whether the selection of the current basis has generated a reward.

\item ${\mathcal {A}}_t$ denotes the action space, that is, the set of possible actions that the agent can select at the current state of the proof process.

\item $r_t$ denotes the immediate reward, which provides feedback to the agent based on the progress toward the proof objective, thereby guiding the agent's learning.
\end{itemize}

We now provide a detailed explanation of the main steps outlined in the {\em Environment Construction} component of Fig.\ref{figure:frame} as follows:

{\bf [Memory]}
In the process of searching for a non-negative representation of inequality, it is essential to maintain a repository of existing axioms or lemma bases that guide the selection of the next basis.
Let ${\mathcal{M}}_t$ denote the set of non-negative basis ${\bf x}^{\alpha}(1-{\bf x})^{\beta}$ at time step $t$, referred to as memory. The memory includes the set of non-negative polynomials (e.g. axioms $1-x_i$, $x_i>0$), and the intermediate results derived from these axioms through inference rules (e.g., the lemma $x_i(1-x_i)>0$).

At time step $t-1$, the DQN agent selects an action $a_{t-1}$ from the available action space ${\mathcal {A}}_{t-1}$ based on the current state. 
This action is then added to the memory $\mathcal {M}_{t-1}$, thereby updating $\mathcal {M}_{t}$, as follows:
$${\mathcal {M}}_{t} = {\mathcal {M}}_{t-1}\cup \{a_{t-1}\}.$$
Then, we solve the following optimization problem to determine whether the currently selected basis contributes to obtaining a non-negative representation,
\begin{equation}\label{opt:memaryLP}
\left.\begin{array}{l}
{\rm max}\quad \gamma \\
s.t. \quad f({\bf x})-\gamma =\sum_{i=1}^{|{\mathcal {M}}_t|}\lambda_im_i, \\
\quad\quad\;\lambda \geq 0,
\end{array}\right\}
\end{equation}
where $m_i \in {\mathcal{M}}_t$. 
This problem is reformulated as the LP problem (\ref{opt:matrix}) to obtain the solution $\gamma^{*}$.

In practice, the initial memory $\mathcal{M}_0$ can be defined as follows:
$$\mathcal{M}_0=\{{\bf x}^{\alpha}(1-{\bf x})^{\beta}||\alpha|+|\beta|\leq k\},$$
where $k$ is a non-negative integer determined according to the specific problem. Here, we set $k = \deg(f({\bf x}))$, ensuring that the LP problem (\ref{opt:matrix}) derived from (\ref{opt:memaryLP}) has a feasible solution.

{\bf [Reward Design]}
After executing the action $a_t$, the corresponding immediate reward $r_t$ from the environment feedback will be obtained. The reward measures the feedback obtained by the agent after performing an action in a specific state. The role of the reward is to guide the agent's learning toward the goal or expected behavior.

We employ an unsupervised reward scheme. If the chosen action $a_t$ leads to an improvement in $\gamma_t$, we give a positive reward, otherwise a negative reward.
We design the following reward function to measure how close we are to achieving the proof goal based on the relative improvement of the boundary:

\begin{equation}
r_t = 
\begin{cases}
\frac{\gamma_{t}-\gamma_{t-1}}{|\gamma_0|} & \text{if } \gamma_t \neq \gamma_{t-1} \\
\epsilon & \text{if } \gamma_t = \gamma_{t-1}
\end{cases}
\end{equation}

where $\gamma_t$ represents the solution to the LP problem (\ref{opt:matrix}) obtained using the Gurobi solver \cite{gurobi} at step $t$, $\gamma_0$ denotes the initial value computed based on 
${\mathcal {M}}_0$, and $\epsilon$ represents a small negative constant. When the boundary value improves, the reward is normalized by dividing it by 
$|\gamma_0|$; this normalization promotes the convergence and stability of the training process. In cases where the boundary value does not improve, a small penalty is imposed to incentivize the agent to select actions that lead to boundary improvement.

{\bf [State]}
From the above analysis, it can be observed that the solution $\gamma_t$ to the LP problem (\ref{opt:matrix}) provides crucial information for evaluating the effectiveness of the selected action $a_t$.
It should be noted that if the current action $a_t$ does not result in a reward, it does not necessarily mean that it is an invalid action, it may still affect the selection of subsequent actions.

Therefore, we represent the state $s_t$ as a tuple: $s_t=[\gamma_t,\kappa]$. The first element is the optimal solution of (\ref{opt:matrix}) at the current time step $t$, while the second element $\kappa$ records the number of consecutive rounds in which the action has not yielded a reward.

{\bf [Action Space]}
At each time step $t$, all the elements in ${\mathcal {M}}_{t}$ are verified to be non-negative through rule (\ref{proveruel}). We can construct the action space ${\mathcal {A}}_t$ for the current time step based on ${\mathcal {M}}_{t}$. 
Specifically, we select an element $m$ from ${\mathcal {M}}_{t}$, and choose to multiply it by $x_i$ or $1-x_i$, thus obtaining $2n|{\mathcal {M}}_{t}|$ action options.  

In practice, it is not necessary to calculate ${\mathcal {A}}_t$ every time, we adopt an incremental update method, namely:
$$\mathcal {A}_t=\mathcal {A}_{t-1} \cup \{a_{t-1} x_i, a_{t-1}(1-x_i)| i=1,\ldots,n\},$$
where $a_{t-1}$ is the action chosen by the agent at time step $t-1$.At the same time, we observed that the action space is gradually increasing with time steps. In other words, the action space is variable, which is why we chose the DQN algorithm.

\subsection{DQN Training}
The goal of reinforcement learning is typically to maximize the expected cumulative return $G_t =
\mathbb{E}[\sum_{t=0}^{T}\delta^{t}r_{t}]$, where $T$ is the terminal time step, $\delta$ is the discount factor, and $r_t$ is the immediate reward at time step $t$.
The role of the discount factor is to balance the relationship between current rewards and future rewards, with its value constrained within the range of 0 and 1.

In order to solve the above reinforcement learning problem, we use the DQN algorithm. There are two neural networks in the DQN algorithm, namely the Q-value neural network $Q_{\theta}(s,a)$ and the target Q-value neural network $Q_{\theta'}(s,a)$.
This neural network $Q_{\theta}(s,a)$ takes as input the current state $s$ and a candidate action $a$ and outputs an estimate of the expected future return for that state-action pair, where $\theta$ represents the trainable parameters of the network.
The parameters $\theta'$ in the target network $Q_{\theta'}(s,a)$ are updated every $c$ step, assigning the parameters $\theta$ to $\theta'$ to enhance the stability of the network, where $c$ is a hyperparameter.

The DQN algorithm has proven effective in learning robust decision-making policies in complex, high-dimensional environments.
At each time step t, we can put the generated quadruple $(s_t, a_t, r_t, s_{t+1})$ into the replay buffer $R$.
We sample $N$ data ${\{(s_i,a_i,r_i,s_{i+1})\}}_{i=1,\dots,N}$ from the replay buffer $R$ and update the Q-value network $Q_{\theta}(s,a)$ by minimizing the loss function:
\begin{equation}\label{lossFUN}
 L=\frac{1}{N}\sum_{i=1}^{N}(r_i+\delta\max_{a'}Q_{\theta^{'}}(s_{i+1},a')-Q_{\theta}(s_i,a_i))^2,
\end{equation}
where $\delta$ is a discount factor. We use the $AdamW$ optimizer to update the parameters $\theta$.

The DQN agent selects an action $a$ using the $\epsilon$-greedy strategy when in state $s$:
\begin{equation*}
\pi(a|s) = 
\begin{cases}
1 - \epsilon + \frac{\epsilon}{|A(s)|} & \text{if } a = \arg\max_a Q(s,a) \\
\frac{\epsilon}{|A(s)|} & \text{if } a \neq \arg\max_a Q(s,a)
\end{cases}
\end{equation*}
where $\epsilon$ is a hyperparameter, $|A(s)|$ is the size of the action space in state $s$, and $\pi$ is the policy.


\subsection{Fast Multivariate Polynomial Multiplication}
We propose a fast method for multivariate polynomial multiplication to speed up the polynomial multiplication computations involved in executing actions.
Given two polynomials $p({\bf x}),q({\bf x})\in \mathbb{R}[{\bf x}]$, the product $p({\bf x})\cdot q({\bf x})$ can be computed efficiently through the following steps: 

\noindent
{\bf (i) Transforming the multivariate polynomials into univariate polynomials:} For the given pair of polynomials:
\begin{equation}
\begin{array}{ll}
 p({\bf x})=a_1{\bf x}^{\alpha_1}+a_2{\bf x}^{\alpha_2}+\cdots+a_r{\bf x}^{\alpha_r},\\ \nonumber
 q({\bf x})=b_1{\bf x}^{\beta_1}+b_2{\bf x}^{\beta_2}+\cdots+b_s{\bf x}^{\beta_s},
\end{array}
\end{equation}
where $\alpha_i,\beta_j\in \mathbb{N}^n$, we set an upper bound $D > d_p \cdot d_q$ for the degree of $p({\bf x})\cdot q({\bf x})$ to ensure effective transformation. Then the multivariate polynomial $p({\bf x})$ can be transformed into the univariate polynomial 
$\tilde{p}(z) = \sum_{i=1}^ra_iz^{\nu_i}$, 
where
\begin{equation}\label{univariate}
\nu_i = \sum_{k=1}^n\alpha_{i_k}D^{k-1},\,i=1,2,\ldots,r.
\end{equation}
Here, $(\alpha_{1_i}, \ldots, \alpha_{n_i})$ are the exponents associated with the monomial $\prod_{k=1}^nx_k^{\alpha_{i_k}}$.
Similarly, we transform the multivariate polynomial $q({\bf x})$ into the univariate polynomial $\tilde{q}(z)=\sum_{j=1}^sb_jz^{\kappa_j}$. 


{\bf (ii) FFT-based univariate polynomial multiplication:}
At this stage, we compute the product of $\tilde{p}(z)$ and $\tilde{q}(z)$, which is equivalent to computing the coefficients of the polynomial $g(z)=\sum_{i=0}^{{\nu_r}+{\kappa_s}}c_iz^i=\tilde{p}(z)\cdot\tilde{q}(z)$.
The coefficient vector $(c_0,\ldots,c_{\nu_r+\kappa_s})$ of the polynomial $g(z)$ is also commonly referred to as the convolution of the coefficient vectors $(a_1,\ldots,a_r)$ and $(b_1,\ldots,b_s)$ of the polynomials $\tilde{p}(z)$ and $\tilde{q}(z)$, respectively. To accelerate polynomial multiplication, we utilize the FFT method as follows.
\begin{itemize}
\item calculate the values $A_j=a_1\omega_{N}^{j\nu_1}+\cdots+a_r\omega_{N}^{j\nu_r}$ and $B_j=b_1\omega_{N}^{j\kappa_1}+\cdots+b_s\omega_{N}^{j\kappa_s}$, $j=0,1,\ldots N-1$;

\item compute the product $C_j=A_jB_j$, $j=0,1,\ldots,N-1$;

\item calculate and output the values $c_k=\frac{1}{N}\sum_{j=0}^{N-1}C_j\cdot e^{i2\pi \frac{kj}{N}}$, here $N=2^\theta$, $\theta =\lceil \log_2(\nu_r+\kappa_s+1)\rceil$.

\end{itemize}

Following this method, we obtain the product $\tilde{p}(z)\cdot \tilde{q}(z)$, denote as 
$g(z)=c_1z^{d_1}+c_2z^d_2+\cdots+c_tz^{d_t}.$

{\bf (iii) Mapping back to the multivariate representation:}
Finally, the resulting univariate polynomial $g(z)$ 
is mapped back to a multivariate form to obtain
the target multivariate polynomials. Specifically, each term $z^{d_i}$ ($i=1,2,\ldots,t$) is replaced with the expression $x_1^{\mu_{i_1}} x_2^{\mu_{i_2}}\cdots x_n^{\mu_{i_n}}$, where the exponents $(\mu_{i_1}, \mu_{i_2}, \ldots, \mu_{i_n})$ are determined by the following Euclidean algorithm:
\vspace{-1ex}
\begin{equation}\label{eq:CRT}
\left\{
\begin{array}{ll}
\mu_{i_1}= d_i\; {\rm mod}\; D, \; \rho_1=(d_i-\mu_{i_1})/ D, \\ 
\mu_{i_2}=\rho_1 \;{\rm mod}\; D,\; \rho_2=(\rho_1-\mu_{i_2})/ D,\\
\quad\quad \vdots\\
\mu_{i_n}=\rho_{n-1} \; {\rm mod}\; D.
\end{array}
\right.
\end{equation}
This allows us to recover the original multivariate structure of the polynomial $p({\bf x})q({\bf x})$ from the transformed univariate representation $g(z)$.

In summary, we have discussed the transformation of the polynomial inequality proof problem into a linear programming problem, the construction of a reinforcement learning framework to train an agent to dynamically prove the inequality.  
In addition, we propose a fast multivariate polynomial multiplication method based on multivariate affine transformation and FFT to accelerate the training process. The detailed procedures are summarized in Algorithm \ref{algo1}.

\begin{algorithm}[h]
\caption{APPIRL: Automated Proof of Polynomial Inequalities via Reinforcement Learning}
\label{algo1}
\textbf{Input}: A polynomial $f({\bf x})$; $\mathcal {S}=\{{\bf x}\in\mathbb{R}^n|x_i\in[a_i,b_i] \}$;

\quad \quad\quad Maximum time step $Maxstep$; an episode $T$\\
\textbf{Output}: $bSet$

\begin{algorithmic}[1] 
\STATE Regularize $\tilde{{\bf x}} \in [0, 1]^n$, and obtain $f(\tilde{{\bf x}})$;
 

\STATE  Initialize $bSet\leftarrow$ FALSE, $k=0$ and memory $\mathcal {M}$; 

\STATE Initialize $Q_{\theta}(s,a)$ and $Q_{\theta'}(s,a)$; 


\STATE $\gamma\leftarrow SolveLP(f({\bf x}),\mathcal {M})$;

\STATE {\bf for } $e=1,\ldots,T$ {\bf do}
 \STATE \quad Obtain the initial state $s$;  
\STATE \quad {\bf while} {$\gamma<0 \wedge k\leq Maxstep$}  
\STATE \quad\quad $a\leftarrow Agent(Q(s,\mathcal{A}))$; 

\STATE \quad\quad Update $\mathcal{M} \leftarrow \mathcal {M}\cup \{a\}$;

\STATE \quad\quad $\gamma\leftarrow SolveLP(f({\bf x}),\mathcal {M})$;

\STATE \quad\quad Update next state $s'$ and reward $r$;

\STATE \quad\quad$\mathcal {A}\leftarrow FastPoly.Mult.(\mathcal{M},a)$;

\STATE \quad\quad Store $(s, a, r, s')$ in $R$;

\STATE \quad\quad Sample $\{(s_i, a_i, r_i, s'_{i+1})\}_{i=1,\ldots,N}$ from $R$;

\STATE \quad\quad Update $Q_{\theta}$ by minimizing (\ref{lossFUN});
\STATE \quad\quad $k=k+1$;

\STATE \quad {\bf end}

\STATE \quad $\gamma\leftarrow  SimulationTrained\_Q$;

\STATE \quad{\bf if} {$\gamma\geq 0$} {\bf then}

\STATE \quad\quad $bSet \leftarrow$ TRUE;
\STATE\quad{{\bf end if}}
\STATE {\bf end}
    
\STATE \textbf{return} $bSet$
\end{algorithmic}
\end{algorithm}

The Algorithm \ref{algo1} describes the process of using the DQN method to train an agent for selecting basis.
In this process, both the state representation $s$ and the action $a$ are executed in vector form during the training phase.
Concretely, the implementation process consists of four main steps:

(i) { Preprocessing} (Lines 1-6):
Regularizing the variables ${\bf x}$ to preprocess the inequality $f({\bf x})$. The proof state, environmental variables, and the $Q$-network are initialized. 

(ii) { Environment Construction} (Lines 8-15): 
The agent selects and executes $a$, updating the memory $\mathcal {M}$, obtain the
reward $r$ and make the state change. The action space $\mathcal {A}$ is constructed using the fast multivariate polynomial multiplication to accelerate the construction of basis.

(iii) { Loss minimizing and training} (Lines 16-17): The target loss is minimized to update the network parameters, which enables the training of the agent to optimally select proof strategies.

(iv) { Simulation} (Lines 20-23): 
Simulating the trained agent to obtain the proof state and determine whether the polynomial inequality holds.

\section{Experiments}\label{ex}

In this section, we evaluate the effectiveness of our method by comparing it with a random search strategy.
In addition, we compared our method with LDPP \cite{FawziMFF19} and S2V-DQN \cite{wang2021solving} on the maximum stable set problem in graphs, and the results demonstrate the effectiveness of our approach.
All experiments were conducted on a machine with Ubuntu 24.10, 1TB RAM, dual Intel(R) Xeon(R) Gold 6430 CPUs, and two NVIDIA L40S GPUs.
The code is available at \url{https://github.com/blliu6/APPIRL}.

\subsection{Performance Evaluation}\label{subsec:benchmark}

We implemented a dynamic inequality proof tool called "APPIRL" based on Algorithm \ref{algo1}. We conducted experimental evaluations on a set of general inequality proof benchmarks to demonstrate the effectiveness of our method compared to a {\em Random Search} strategy. 

Table \ref{table2} records the best results obtained from exploring different network architectures.
For the {\em Random Search} method, we randomly select a strategy for proving at each step, and perform $50$ trials for each instance to minimize randomness. To manage computational efficiency, we set an initial bound according to $\deg(f({\bf x}))$ when constructing $\mathcal{M}_0$.

\begin{table}[ht]
\renewcommand\arraystretch{1}
\caption{Comparison of the Performance Evaluation}\label{table2}
\centerline{
\scalebox{1}{
\begin{tabular}%
{
@{\vbox to 2.0ex{\hbox to 0.2em{\hss}\vss}}l@{\hbox to 0.2em{\hss}} 
@{\hbox to 0.1em{\hss}}c@{\hbox to 0.1em{\hss}}
@{\hbox to 0.1em{\hss}}c@{\hbox to 0.1em{\hss}}|
@{\hbox to 0.0em{\hss}}c@{\hbox to 0.2em{\hss}} 
@{\hbox to 0.1em{\hss}}r@{\hbox to 0.1em{\hss}}
@{\hbox to 0.1em{\hss}}r@{\hbox to 0.2em{\hss}}|
@{\hbox to 0.1em{\hss}}r@{\hbox to 0.0em{\hss}}
@{\hbox to 0.0em{\hss}}r@{\hbox to 0.0em{\hss}}
@{\hbox to 0.0em{\hss}}r@{\hbox to 0.2em{\hss}}
}
\toprule[1pt]
\multirow{2}{*}{\textbf{Ex.}}
&\multirow{2}{*}{\textbf{$n_{\bf x}$}}
&\multirow{2}{*}{\textbf{$d_f$}}
&\multicolumn{3}{@{\hbox to 0.2em{\hss}}c|@{\hbox to 0.1em{\hss}}}{{\textbf{APPIRL}}}
&\multicolumn{3}{@{\hbox to 0.1em{\hss}}c@{\hbox to 0.1em{\hss}}}{{\textbf{Random Search}}}\\ \cline{4-9} 
& & &\;NN &$\mathcal{M}_0$ &$S$\; &\;$S_{max}$ & \;\;$S_{min}$ &\;\;$S_{avg}$
  \\  \hline
$C_1$\cite{verschelde1996polynomial} &2 &2  &\;64(4) &14 &{\bf 7} &81\; &8\;\; &31 \\
  
$C_2$\cite{verschelde1996polynomial} &2 &2  &\;64(4) &14 &{\bf 19} &90\; &13\;\; &43 \\
  
$C_3$\cite{iserles1995personal} &3 &2  &\;96(4) &27 &{\bf 17} &268\; &7\;\; &76 \\

$C_4$\cite{1989neural} & 3 & 3  &\;128(4)&83 &{\bf 25} &382\; &37\;\;  & 148\\

$C_5$\cite{verschelde2001phc} &4 &3 &\;128(4) &164  &{\bf 8} &806\; &14\;\;  &238\\

$C_6$\cite{1989neural} &4 &3  &\;128(4) &164  &{\bf 23} &689\; &28\;\;  &211\\

$C_7$\cite{faugere1999new} &4 &4  &\;160(4) &494  &{\bf 391} &1754\; &491\;\;  &859\\

$C_8$\cite{wright1985finding} &5  &2  &\;96(4)  &65  &{\bf 3}  &1049\; &51\;\; &353 \\

$C_9$\cite{emiris1994sparse} &5 &4  &\;128(4)&1000  &{\bf 159} &2980\; &516\;\; &2121\\

$C_{10}$\cite{emiris1997general} &6 &2 &\;160(4) &90 &{\bf 102} &627\;  &159\;\;  &371 \\
\bottomrule[1pt]
\end{tabular}}
}\par
\end{table}

In Table \ref{table2}, the first column lists the origins of the $10$ widely used examples.
The symbol $n_{\bf x}$ represents the number of dimensions of the variables in the given polynomial inequality, and $d_f$ denotes the maximum degree of the polynomial.
'NN' represents the network architecture used. For example, the network for $C_1$ is trained with $4$ hidden layers, each comprising $64$ neurons, using ReLU activation functions. The column '$\mathcal{M}_0$' denotes the initial memory size, and $S$ indicates the number of time steps required to complete the proof using our method. 
The '{\em Random Search}' column records the maximum $S_{max}$, minimum $S_{min}$, and the average $S_{avg}$ time steps.

From Table \ref{table2}, we can observe that {\em APPIRL} consistently outperforms the {\em Random Search} method in terms of the number of steps required to complete the proofs. 
For all examples except $C_2$ and $C_3$, 
{\em APPIRL} requires fewer steps than even the minimum steps $S_{min}$ of the {\em Random Search} method.
Especially, for $C_8$, {\em APPIRL} only requires $3$ steps, while {\em Random Search} requires $51$ steps, is $17$ times more than {\em APPIRL} at $S_{min}$ and $117$ times more on $S_{avg}$. Across the $10$ examples in the table, {\em APPIRL} requires an average of $75.4$ steps, while {\em Random search} requires an average of $445.1$ steps, approximately $6$ times higher than {\em APPIRL}.

For problems with the same dimensionality but different degrees, such as $C_3$ and $C_4$, $C_6$ and $C_7$, etc, {\em APPIRL} shows an increase in both memory size and number of steps as the degree of the polynomial increases. This is likely due to the higher complexity of higher-degree polynomials, which expands the possible action space and requires more computational resources. Interestingly, for $C_5$ and $C_6$ having the same dimensionality and degree, the required time steps differ, which is due to inherent variability in reinforcement learning. 
In contrast, the {\em Random Search} method has a more consistent performance across the examples, as it is mainly influenced by the size of the search space.

These observations demonstrate that our method ({\em APPIRL}) achieves efficient dynamic proving with fewer time steps for various polynomial inequality problems while maintaining good expressiveness. This adaptive performance highlights the effectiveness of combining FFT-based fast polynomial multiplication with dynamic proving through deep learning. Compared to traditional search methods, our approach offers a significant advantage of greater efficiency and precision, especially when handling complex mathematical tasks. 

\subsection{Application: Maximum stable sets in graphs}\label{subsec:Stableset}

We now consider a combinatorial optimization problem using our dynamic proof method. Consider an undirected graph $\mathcal {G} = (\mathcal {V},\mathcal {E})$, where $\mathcal {V}=\{1,2,\ldots,n\}$ represents the set of nodes and $\mathcal {E} \subseteq \mathcal {V}\times \mathcal {V}$ represents the set of edges. A stable set $S \subseteq \mathcal{V}$ is a set of vertices with no two vertices connected by an edge \cite{gomory1961multi}. 

Finding a stable set with maximum cardinality can be formally stated as:
\begin{equation}\label{Ex.State}
\left.\begin{array}{ll}
{\max} \quad \sum_{i=1}^nx_i  \\
s.t. \quad\;\; x_ix_j=0,\quad \forall (i,j)\in \mathcal {E}, \\
\quad\quad \quad x_i^2=x_i, \quad \forall i\in \mathcal {V}.
\end{array}\right\}
\end{equation}
The constraint $x_ix_j=0$ ensures that no two vertices in the stable set $S$ are connected by an edge, while $x_i^2=x_i$ enforces $x_i\in \{0,1\}$. To verify that no stable set exceeds the size $|S|$, we need to prove $|S| -\sum_{i=1}^nx_i\geq 0$.
The problem of finding a stable set of maximum cardinality is a classic NP-hard problem with numerous applications across diverse fields \cite{athuOP13,ConfortiR04}. Now, we illustrate our dynamic proof approach using this problem as an example.

\begin{figure}[h]
  \centering {
\includegraphics[width=0.29\textwidth]{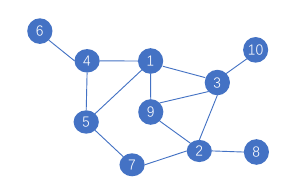}
   }\caption{ An undirected graph $\mathcal {G} = (\mathcal {V},\mathcal {E})$,  The vertices set is $\mathcal {V}=\{x_1,\ldots,x_{10}\}$, and the edge set is $\mathcal {E} = \{x_ix_j = 0,x_i,x_j\in \mathcal {V}\}$, where $x_i$ and $x_j$ are vertices connected by an edge.} \label{fig1}
\vspace{0ex}
 \end{figure}

Consider the undirected graph depicted in Fig. \ref{fig1}. In this graph, the maximum stable set is $|S|=5$, we choose the maximum stable set as $S=\{x_5, x_6, x_8, x_9, x_{10}\}$. Therefore, the goal is to prove the inequality $5-\sum_{i=1}^{10}x_i\geq 0$.
 
\vspace{0ex}
 
\begin{figure}[h]
  \centering {  \includegraphics[width=0.4\textwidth]{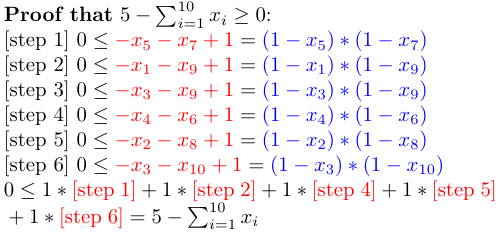}
   }
   \caption{An example of proof generated by our agent} \label{fig2}
 \end{figure}

We limit the dynamic proof to a maximum of $100$ steps and restrict the degree of intermediate lemmas to $2$. Using Algorithm \ref{algo1}, the inequality is proven in just $6$ steps, as shown in Fig. \ref{fig2}.
 
In Fig. \ref{fig2}, the axioms are represented in blue, while the derived polynomials (i.e., intermediate lemmas) are shown in red. It's important to note that all the coefficients in the proof are rational numbers. This ensures that the proof is not only exact but also fully verifiable.

To fully demonstrate the feasibility of our proposed method, we conducted a comparative analysis against the approach presented in \cite{FawziMFF19} on a set of benchmark instances for the maximum stable set problem, as shown in Table \ref{table1}. The {\em LDPP} method aims to construct a deep reinforcement learning framework that leverages the inherent symmetry of the problem as inductive bias, learning polynomial embeddings and guiding the selection of inference rules.

In Table \ref{table1}, the first column lists the examples which are taken from the complete set of experiments in {\em}\cite{FawziMFF19}, where $v_{\bf x}$ represents the number of vertices in the graph. The size of the maximum stable set is denoted as "mss".
The last column records the number of time steps required by the {\em LDPP} method to complete the proof. The code of the {\em LDPP} method is not open source, the data in this column is taken from \cite{FawziMFF19}.
\begin{table}[ht]
\renewcommand\arraystretch{1}
\caption{Performance Evaluation compared to {\em LDPP} on the maximum stable set problem.}\label{table1}
\centerline{
\scalebox{1}{
\begin{tabular}%
{
@{\vbox to 2.0ex{\hbox to 0.5em{\hss}\vss}}l@{\hbox to 0.5em{\hss}} 
@{\hbox to 0.6em{\hss}}c@{\hbox to 0.6em{\hss}}
@{\hbox to 0.6em{\hss}}c@{\hbox to 0.2em{\hss}}|
@{\hbox to 0.0em{\hss}}c@{\hbox to 0.4em{\hss}} 
@{\hbox to 0.4em{\hss}}c@{\hbox to 0.6em{\hss}}|
@{\hbox to 0.0em{\hss}}c@{\hbox to 0.6em{\hss}}
}
\toprule[1pt]
\multirow{2}{*}{\textbf{Ex.}} 
&\multirow{2}{*}{$v_{\bf x}$} 
&\multirow{2}{*}{$mss$\;} 
& \multicolumn{2}{@{\hbox to 0.2em{\hss}}c|@{\hbox to 0.0em{\hss}}}{{\textbf{APPIRL}}}
& \multirow{2}{*}{\;\;\;\textbf{LDPP} \cite{FawziMFF19}} \\ \cline{4-5} 
&  &   &\;\;\;$\mathcal{M}_0$  &$S$  & \\ \hline


$G_1$&7  & 1    &14  &6  &6   \\

$G_2$&10  &5    &20  &6  &6    \\

$G_3$&10  &4   &20  &37 &42    \\ 






\bottomrule[1pt]
\end{tabular}}
}\par
\end{table}

Table \ref{table1} presents a comparison of dynamic proof results for three sets of classic maximum stable set verification problems from \cite{FawziMFF19}. 
Observing the data in the table, we find that for $G_1$ and $G_2$, the number of time steps required by APPIRL matches that of LDPP, demonstrating that this number of steps represents the optimal automated proof path. 
For $G_3$, our method completes the proof in just 37 steps, while LDPP requires 42 steps. 
Although APPIRL is designed for general inequality proof problems and LDPP leverages the inherent symmetry of stable set problems, our method still outperforms LDPP.

\begin{table}[h]
\renewcommand\arraystretch{1}
\caption{Performance Evaluation compared to {\em S2V-DQN} on the maximum stable set problem.}\label{table3}
\centerline{
\scalebox{1}{
\begin{tabular}{
@{\vbox to 2.0ex{\hbox to 0.5em{\hss}\vss}}c@{\hbox to 0.5em{\hss}}| 
@{\hbox to 0.5em{\hss}}c@{\hbox to 0.5em{\hss}}
@{\hbox to 0.5em{\hss}}c@{\hbox to 0.5em{\hss}}
@{\hbox to 0.3em{\hss}}c@{\hbox to 0.3em{\hss}}
@{\hbox to 0.3em{\hss}}c@{\hbox to 0.3em{\hss}} 
@{\hbox to 0.3em{\hss}}c@{\hbox to 0.3em{\hss}}
@{\hbox to 0.3em{\hss}}c@{\hbox to 0.3em{\hss}}
@{\hbox to 0.3em{\hss}}c@{\hbox to 0.3em{\hss}}
@{\hbox to 0.3em{\hss}}c@{\hbox to 0.3em{\hss}}
}
\toprule[1pt]
\textbf{Ex.} &$G_1$ &$G_2$ & $G_3$  &$R_1$  &$R_2$ &$R_3$ &$R_4$ &$R_5$ \\ \hline

$\textbf{n}$ & 7 &10 &10 &15 &15 &20 &20 & 20\\ \hline 

$\textbf{v}$ &21 &13 &15 &75 &82 &133 &138 &139\\ \hline

$\textbf{S2V-DQN}$ &1 &5 &4 &7 &8 &7 & 9 &10\\ \hline

$\textbf{APPIPL}$ & 1 &5 &4 &7 &8 &7 &9 &10\\

\bottomrule[1pt]
\end{tabular}}
}\par
\end{table}

To further verify the performance of our method on the maximum stable set problem, we conducted a set of comparative experiments with the S2V-DQN method proposed in \cite{wang2021solving}, which are shown in Table \ref{table3}.
We added graph $R_1-R_5$ in addition to graph $G_1$ to $G_3$ as new evaluation benchmarks.
Among them, $n$ is the number of vertices in the graph, $v$ is the number of edges in the graph. 
On the problem of the maximum stable set, the size of the maximum stable set found by our method LDPP is the same as that found by S2V-DQN, yielding the same results. 
Our method is not only applicable to general inequality proving problems, but also shows excellent performance on the maximum stable set problem.
\section{Conclusion}
\label{sec:con}
We proposed a novel RL-based method for proving polynomial inequalities. Using the Positivstellensatz theorem, we formulated the inequality-proving problem as obtaining a non-negative representation in the Krivine-basis form. We used RL to train an agent to gradually select an effective basis, minimizing the number of proof rounds and resulting in a representation. And we innovatively incorporated LP solving, multivariate affine transformation and FFT into the reinforcement learning process to improve the efficiency of inequality proving. Notably, extensive complexity analysis and experimental findings consistently show that the proposed approach is both effective and scalable. Furthermore, our proposed approach can also applicable to attacking the problem of maximum stable set in graphs.

\noindent{\bf{Acknowledgements.}}This work was supported by the GPU computing platform of the Academy of Mathematics and Systems Science, Chinese Academy of Sciences, and in part by the National Key Research and Development Program of China under Grant (2023YFA1009402), the National Natural Science Foundation of China under Grants (No.62472362, 62272397, 12171159), the Innovation Program of Shanghai Municipal Education Commission under Grant (2021-01-07-00-08-E00101), the "Digital Silk Road" Shanghai International Joint Lab of Trustworthy Intelligent Software under Grant (No.22510750100), and the Shanghai Trusted Industry Internet Software Collaborative Innovation Center.

\bibliographystyle{unsrt}
\bibliography{references}

\end{document}